\documentclass[10pt,twocolumn,letterpaper]{article}

\usepackage{cvpr}              

\definecolor{cvprblue}{rgb}{0.21,0.49,0.74}
\usepackage[pagebackref,breaklinks,colorlinks,allcolors=cvprblue]{hyperref}

\def\paperID{*****} 
\def\confName{CVPR}
\def\confYear{2025}

\title{Beyond Correlation: Towards Causal Large Language Model Agents in Biomedicine}

\author{Adib Bazgir\\
University of Missouri-Columbia\\
Columbia, MO 65211, USA\\
\and
Amir Habibdoust Lafmajani\\
University of Missouri-Columbia\\
Columbia, MO 65211, USA\\
\and
Yuwen Zhang \thanks{Corresponding Author: Yuwen Zhang at zhangyu@missouri.edu.}\\
University of Missouri-Columbia\\
Columbia, MO 65211, USA\\
}

\begin{document}
\maketitle
\begin{abstract}
Large Language Models (LLMs) show promise in biomedicine but lack true causal understanding, relying instead on correlations. This paper envisions causal LLM agents that integrate multimodal data (text, images, genomics, etc.) and perform intervention-based reasoning to infer cause-and-effect. Addressing this requires overcoming key challenges: designing safe, controllable agentic frameworks; developing rigorous benchmarks for causal evaluation; integrating heterogeneous data sources; and synergistically combining LLMs with structured knowledge (KGs) and formal causal inference tools. Such agents could unlock transformative opportunities, including accelerating drug discovery through automated hypothesis generation and simulation, enabling personalized medicine through patient-specific causal models. This research agenda aims to foster interdisciplinary efforts, bridging causal concepts and foundation models to develop reliable AI partners for biomedical progress.
\end{abstract}    
\section{Introduction}
\label{sec:intro}

Large Language Models (LLMs) have demonstrated unprecedented flexibility across many tasks in medicine, from answering clinical questions to interpreting biomedical literature \cite{author1}. However, these models primarily learn correlations in data rather than true causal relationships \cite{author3}. Correlation is not causation – a mantra especially critical in biomedical research \cite{author2}where understanding why something works is often more important than what correlates with an outcome. Standalone LLMs hallucinate, rely on outdated data, and lack causal understanding, limiting their use in medicine \cite{author4}. While LLMs excel at processing text and capturing domain knowledge implicitly, even generating plausible causal arguments \cite{author5}, distinguishing cause from effect or performing robust counterfactual reasoning remains a challenge. They may exhibit 'causal blindness', struggling to infer causality directly from complex biological data \cite{author6}. Meanwhile, multimodal foundation models are emerging that can interpret combinations of text, imaging, genomics, and other data, producing rich outputs like free-text explanations or image annotations \cite{author7}. The convergence of multi-domain data and LLM capabilities presents a new opportunity to create causality-aware LLM agents for biomedicine. Such agents could integrate multimodal biomedical data and reason like scientists, envisioning interventions and predicting outcomes. Recent work has explored using LLMs within agentic frameworks for hypothesis generation \cite{author8}, interacting with tools \cite{author9,author10}, structuring unstructured real-world data (RWD) like clinical notes \cite{author11}, automating causal discovery workflows \cite{author9}, converting narrative text into analyzable graph structures \cite{author12}, and even engaging users conversationally \cite{author13}.
However, achieving true personalization and reliability requires moving beyond the capabilities of LLMs alone. Integrating structured knowledge sources like knowledge graphs (KGs) offers a path to ground LLM reasoning and enhance explainability, though KGs themselves can be static \cite{author4}. Furthermore, genuine causal understanding necessitates incorporating formal causal discovery and inference methods, such as Mendelian Randomization (MR) or techniques handling confounding in observational data \cite{author9,author11,author6}, potentially applied to individual patient data for personalized insights \cite{author13,author3}. This position paper outlines a vision for causal LLM agents, discussing key challenges to be overcome, including the synergistic integration of LLMs, KGs, causal methods, and multi-modal data, and highlighting potential opportunities across several biomedical domains, such as automated causal knowledge discovery and Real-World Evidence (RWE) generation, as shown in Figure 1.

\section{Challenges in Enabling Causal Reasoning}
\label{sec:formatting}

\subsection{Agentic Framework Design and Control}
Building an agentic LLM framework for biomedicine means allowing the model to autonomously perform tasks such as proposing experiments, retrieving literature, performing analyses, or making clinical suggestions, rather than just passively answering questions. Designing such autonomy raises serious safety and control challenges. In biomedical contexts, an agent’s actions or recommendations can directly impact experiments or patient care, so strict oversight and safeguards are essential. For example, if a causal LLM agent was connected to a lab automation system, a literature database like PubMed, a GWAS data repository like OpenGWAS, or a clinical decision support tool, we must ensure it cannot execute harmful interventions or retrieve and process information incorrectly. Frameworks like MRAgent demonstrate how LLMs can control complex workflows involving multiple external tools and data sources for tasks like automated causal discovery using Mendelian Randomization.
Mechanisms for human-in-the-loop control, as explored in frameworks like MatAgent, permission gating for high-stakes actions, and alignment with ethical guidelines are critical. Furthermore, agents should be bounded in their goals – an “aligned” agent that understands the limits of its authority. Designing a framework where the LLM’s agentic behavior can be audited and constrained is an open challenge. Integrating structured knowledge, such as KGs, can enhance transparency and provide clearer evidence for the agent's decisions, potentially increasing trust. LLMs themselves might be used to convert narrative clinical text, like psychiatric case formulations, into graph-based representations suitable for analysis, although the reliability and validity of such conversions require careful assessment. Researchers have argued that AI agents should be scrutinized like scientists, undergoing rigorous peer review of their plans and reasoning. Achieving the right balance between an agent’s autonomy for efficiency and scale, and reliable control and traceability for safety and trust, will be pivotal for biomedical LLM agents.

\begin{figure}
  \centering
    \includegraphics[width=1\linewidth]{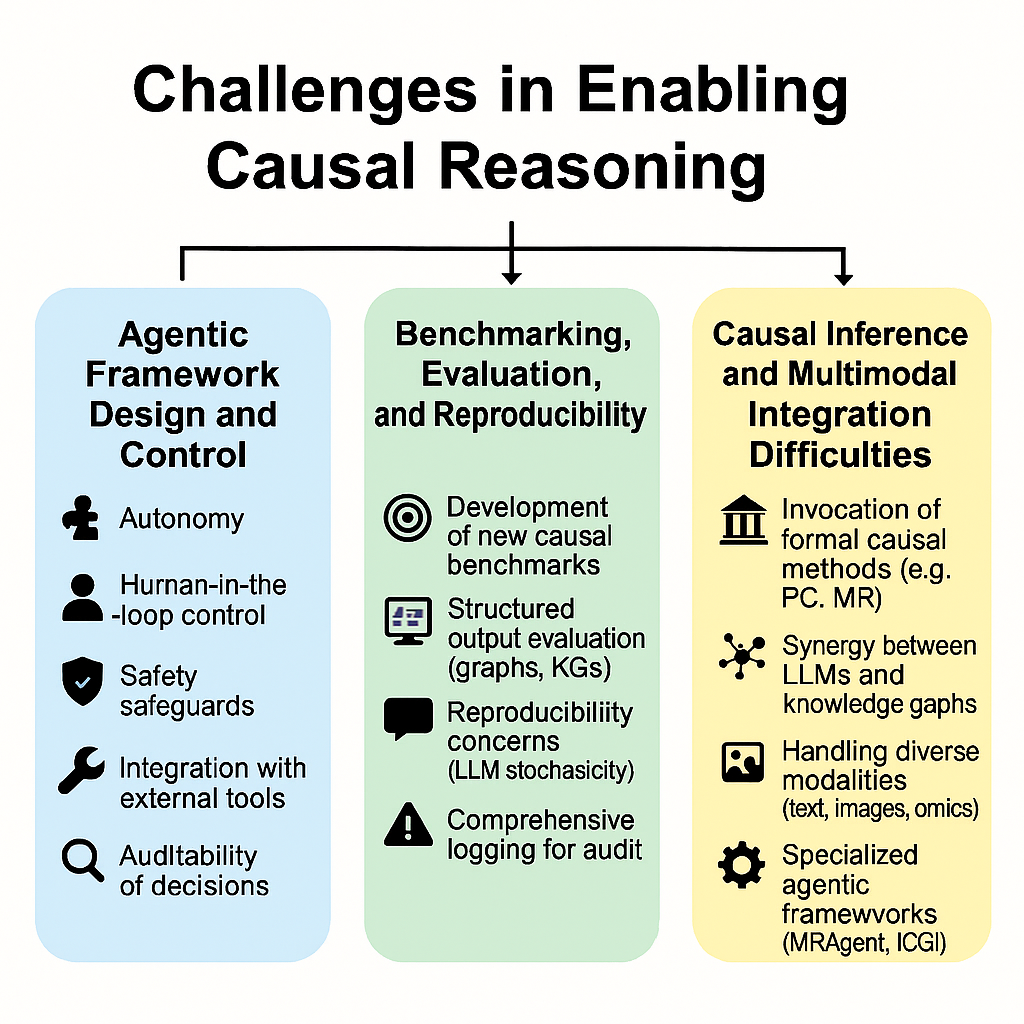}
    \caption{Schematic of major challenges in causal reasoning.}  
  \label{fig:short}
\end{figure}

\subsection{Benchmarking, Evaluation, and Reproducibility}
Evaluating a causal LLM agent’s performance poses unique difficulties. Traditional metrics are insufficient to gauge causal reasoning quality or decision-making safety. New benchmarks and evaluation strategies are needed to test specific causal capabilities. This includes assessing the ability to generate text corresponding to correct causal arguments, reason about counterfactuals, and identify necessary and sufficient causes in vignettes \cite{author5}. Performance on formal causal tasks, like pairwise causal discovery or full graph generation, needs evaluation using established benchmarks (for example, Tübingen pairs) and domain-specific graphs, including tests for generalization to novel datasets created after LLM training cutoffs \cite{author5}.
Alternative evaluation methods are emerging. Structured output evaluation, such as having LLMs generate KGs from medical concepts \cite{author14} or causal graphs from narrative text \cite{author12}, allows comparison against human experts or ground truth using graph similarity metrics (for example, node semantic similarity, Jaccard edge similarity). Task-specific performance evaluation in biomedical applications, like identifying cancer genes \cite{author6} or assessing causality in pharmacovigilance \cite{author15}, often reveals the superiority of domain-specific models. For agentic systems, evaluating the entire workflow execution, comparing against human expert performance on the same multi-step tasks \cite{author9}, provides a holistic assessment. Comprehensive clinical evaluation frameworks, such as CLEVER \cite{author1}, assess LLMs across multiple clinical dimensions (comprehension, reasoning, decision support, risk management) using expert ratings, including performance across different data distributions (ID, OOD, rare diseases) and comparisons against physicians. Multi-agent evaluation, using ensembles or debating frameworks to critique hypotheses \cite{author16,author22}, offers another layer of validation.
However, standardized protocols are lacking, and real-world testing remains limited \cite{author17}. Reproducibility is also a concern due to LLM stochasticity and updates \cite{author18}. While CoT aids auditability \cite{author9,author14}, ensuring verifiable outputs is challenging. Addressing LLM failure modes like hallucination, bias, and sensitivity to prompts requires systematic stress-testing and robust, multi-faceted evaluation methodologies that combine these diverse approaches \cite{author5}.

\subsection{Causal Inference and Multimodal Integration Difficulties}
Current LLMs, even very powerful ones, lack a true grasp of formal causal inference \cite{author5}. While they excel at knowledge-based causal discovery (for example, inferring A→B from variable names) and generating counterfactual arguments from text, they inherently struggle with distinguishing cause from correlation in observational data \cite{author19}, understanding interventions rigorously, and applying formal causal methods without external guidance \cite{author3}. Their textual knowledge does not imply understanding of formal mathematical  and causal logic. Several strategies aim to bridge this gap by integrating LLMs with more formal approaches and diverse data types. One key strategy involves integrating causal tools and methods via agentic frameworks. LLM agents can be designed to invoke external libraries or APIs that implement specific causal algorithms. For instance, the "Causal Agent" framework demonstrates using tools for standard causal discovery (like the PC algorithm) and inference on tabular data \cite{author3}. Domain-specific examples include MRAgent, which automates Mendelian Randomization workflows by calling tools to query PubMed, fetch GWAS data from OpenGWAS, and execute MR analyses using packages like TwoSampleMR \cite{author9}. Similarly, the ICGI framework combines LLM prompting for knowledge retrieval with data-driven causal feature selection using Debiased Machine Learning (DML) on omics data to identify potential cancer genes \cite{author6}.
Another important direction is the synergistic integration of LLMs and Knowledge Graphs (KGs). KGs ground LLM reasoning, improve explainability, and reduce hallucination \cite{author4}. LLMs help build and update KGs by extracting new relationships from text or data. This bi-directional interaction creates a more robust and dynamic knowledge system, addressing the static nature of KGs and the ungrounded nature of LLMs \cite{author4}.
Furthermore, effectively handling multimodal data is crucial for biomedical applications. Agents must process text, images, structured data, and logs. \cite{author10}. LLMs help structure clinical EMR text to enable scalable RWE generation  \cite{author11,author1}. Advanced frameworks like MATMCD explore explicitly integrating multiple modalities, such as using retrieved text or logs alongside statistical causal graphs, employing agents for data augmentation and constraint generation to improve causal discovery \cite{author11}. As discussed, narrative-to-graph conversion is a promising but still unvalidated use case \cite{author12}. Building these capabilities relies heavily on the availability of large-scale, multimodal biomedical datasets for training more capable foundation models \cite{author20}. Crucially, domain-specific pre-training and fine-tuning consistently improve model performance on specialized medical and causal tasks compared to general-purpose models \cite{author15,author1}.

\section{Opportunities and Applications}
As depicted in Figure 2, despite the significant challenges, a future causal LLM agent offers exciting opportunities to transform biomedical research and healthcare. By combining the broad knowledge and pattern recognition of foundation models with principled causal reasoning, such agents could unlock novel applications in several domains.

\begin{figure}
  \centering
    \includegraphics[width=1\linewidth]{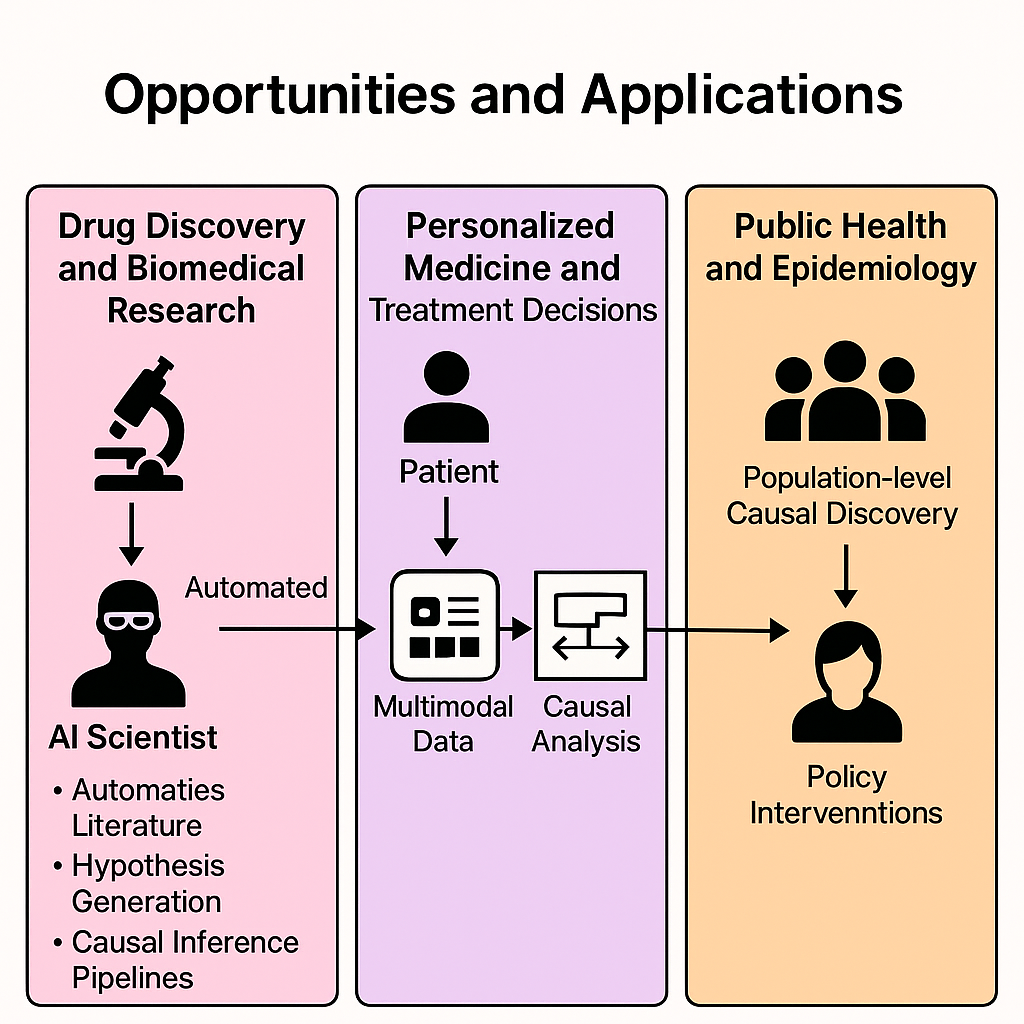}
    \caption{Fully-automated cycle of causal LLM agent workflow in different biomedicine applications.}  
  \label{fig:short}
\end{figure}

\subsection{Drug Discovery and Biomedical Research}
Causally-aware LLM agents can function as "AI scientists," accelerating discovery by automating knowledge synthesis, hypothesis generation, and analysis. A key application is automated causal knowledge discovery. Agents like MRAgent demonstrate the ability to autonomously scan literature, identify potential exposure-outcome pairs for a given disease, check for prior causal analyses (like Mendelian Randomization), retrieve relevant genetic data (OpenGWAS), execute causal inference analyses (using tools like TwoSampleMR), and generate comprehensive reports. This significantly speeds up the process of uncovering potential causal links from existing knowledge and data \cite{author9}.
Building on this, AI-driven hypothesis generation can be enhanced. Agents can synthesize information from literature (potentially via RAG), structured databases, and experimental data to propose testable causal hypotheses. LLMs' strength in generating causal arguments based on their vast training data can bootstrap this process \cite{author8,author16}. 
Furthermore, causal agents can aid in target identification and validation. By integrating LLM-driven causal reasoning with data-driven causal discovery methods applied to omics data (like DML-CGI), potential causal genes or pathways implicated in diseases like cancer can be identified and prioritized, complete with explanations for their proposed roles \cite{author6}. Finally, automated experiment analysis, such as using vision-based agents to detect drug-cell interactions in microscopy images without task-specific training, can streamline experimental workflows. Collectively, these capabilities promise to shorten research cycles, prioritize experiments based on causal plausibility, and uncover novel therapeutic strategies \cite{author21}.

\subsection{Personalized Medicine and Treatment Decisions}
Causality-aware LLM agents can personalize medicine by tailoring decisions to individual causal factors, beyond population averages. A core capability is identifying personal causal drivers. By applying causal discovery and inference techniques, potentially orchestrated by an LLM agent invoking specialized tools, to an individual's longitudinal multi-modal data (from wearables, EHRs, genomics), patient-specific causal links between factors like diet, activity, sleep, and health outcomes can be uncovered. TChatDiet uses N-of-1 analysis to estimate personalized nutrient effects \cite{author13,author3}.
Once personal causal factors are identified (for example, a specific mutation driving a cancer subtype), the agent can tailor interventions and suggest targeted therapies or lifestyle changes addressing the root cause for that individual \cite{author6}. Crucially, LLMs can explain recommendations using patient-specific causal models, improving transparency. This moves beyond simply matching patients to guidelines towards dynamically stratifying them into causal subgroups \cite{author18}.
LLMs also show promise in diagnostic assistance. Advanced medical LLMs like MedFound can generate differential diagnoses and outline reasoning steps, demonstrating effectiveness even for rare diseases \cite{author1}. Integrating causal reasoning capabilities could further enhance diagnostic accuracy by helping to distinguish causes from symptoms or comorbidities. Linking multimodal data by causal relationships can further improve personalized care. 

\subsection{Public Health and Epidemiology}
Automated causal discovery techniques, potentially driven by LLM agents like MRAgent, can systematically scan literature and population-level data (like GWAS) to hypothesize and test potential causal links between environmental exposures, genetic factors, and public health outcomes \cite{author9}. These capabilities can lead to more informed, data-driven public health policies and interventions that are grounded in an understanding of causal mechanisms.

\section{Discussion}
Causal LLM agents are a key step toward reliable, impactful AI in science and biomedicine. This requires moving beyond the correlation-based capabilities of current LLMs towards systems that genuinely understand and manipulate cause-and-effect relationships. Achieving this necessitates a synergistic approach, moving beyond standalone LLMs to integrate their strengths with formal causal methodologies, structured knowledge, and dynamic interaction with external tools. Key research directions emerge from the challenges and opportunities discussed. Firstly, synergistic integration appears paramount. The best path combines LLMs with KGs for grounding and formal tools (e.g., MR, DML) for reasoning. Agentic frameworks are essential for orchestrating these components, enabling LLMs to intelligently invoke external tools, query KGs, and process diverse data inputs.
Secondly, we must effectively leverage LLM strengths while mitigating their weaknesses. LLMs excel at inferring relationships, generating arguments, and extracting causal structures from clinical narratives. These abilities can automate knowledge synthesis and augment human expertise. However, their struggles with formal causal inference from data, susceptibility to hallucination, and lack of inherent grounding necessitate integration with verifiable knowledge sources (KGs) and formal causal methods. Robust evaluation and alignment techniques are crucial for ensuring safety and reliability.
Thirdly, progress relies heavily on the data ecosystem. This includes access to large-scale, diverse, multimodal biomedical datasets and, critically, the ability to structure and integrate Real-World Data (RWD), especially the vast amounts of unstructured text in EMRs. Domain-specific pre-training and fine-tuning of LLMs have proven vital for achieving high performance on specialized medical and causal tasks.
Fourthly, robust evaluation methodologies must be developed and standardized. This goes beyond typical NLP metrics to include specific assessments of causal reasoning, performance across different data distributions and disease rarities, evaluation of agentic task completion against expert benchmarks, and assessment of clinical utility via structured human evaluation frameworks.
Finally, the applications driving this research are transformative. Key opportunities include automating causal discovery from literature and data, generating reliable Real-World Evidence from EMRs, enabling personalized medicine through N-of-1 causal analysis, and providing explainable diagnostic and decision support.

\section{Conclusion}
The development of causal LLM agents for biomedicine requires a shift from standalone models to sophisticated, integrated systems. These agents, empowered by specialized tools, grounded by structured knowledge, and guided by formal causal principles, hold the potential to become invaluable partners for researchers and clinicians. Realizing this vision demands continued interdisciplinary collaboration to tackle the complex challenges of integration, evaluation, validation, and safe, ethical deployment.

\newpage
{
    \small
    \bibliographystyle{ieeenat_fullname}
    \bibliography{main}
}


\end{document}